\documentclass{article} 
\usepackage{times}

\usepackage{amsmath,amsfonts,bm}

\def\eqref#1{equation~\ref{#1}}

\def\1{\bm{1}}

\DeclareMathAlphabet{\mathsfit}{\encodingdefault}{\sfdefault}{m}{sl}
\SetMathAlphabet{\mathsfit}{bold}{\encodingdefault}{\sfdefault}{bx}{n}

\usepackage{hyperref}
\usepackage{graphicx}
\usepackage{subcaption}
\usepackage[super]{nth}
\newcommand{\ts}{\textsuperscript}
\usepackage{algorithm}

\usepackage{algpseudocode}

\usepackage{url}
\usepackage{amsmath,amssymb}
\usepackage{algpseudocode}
\usepackage{upgreek}
\usepackage{pgfplots}
\usepackage{lipsum}
\usepackage{listings}
\usepackage{natbib}

\title{Temporal Shift Reinforcement Learning}

\author{Deepak George Thomas,  
Tichakorn Wongpiromsarn, Ali Jannesari 
\thanks{*This work was not supported by any organization}
\thanks{Deepak George Thomas is with the Department of Computer Science, Iowa State University,
        Ames, IA 50011
        {\tt\small dgthomas@iastate.edu}}%
\thanks{Dr. Ali Jannesari is with the Department of Computer Science, Iowa State University,
        Ames, IA 50011
        {\tt\small jannesar@iastate.edu}}%
\thanks{Tichakorn (Nok) Wongpiromsarn is with the Department of Computer Science, Iowa State University,
        Ames, IA 50011
        {\tt\small nok@iastate.edu}}%
}

\begin{document}

\maketitle

\begin{abstract}
The function approximators employed by traditional image-based Deep Reinforcement Learning (DRL) algorithms usually lack a temporal learning component and instead focus on learning the spatial component. We propose a technique, Temporal Shift Reinforcement Learning (TSRL), wherein both temporal, as well as spatial components are jointly learned. Moreover, TSRL does not require additional parameters to perform temporal learning. We show that TSRL outperforms the commonly used frame stacking heuristic on all of the Atari environments we test on while beating the SOTA for all except one of them. This investigation has implications in the robotics as well as sequential decision-making domains. 
\end{abstract}

\section{Introduction}
Developing Reinforcement Learning (RL) algorithms that can make effective decisions using high dimensional observations such as images is quite challenging. In addition, it consumes a lot of time and energy. In recent months researchers have worked on developing sample efficient RL, plug and play algorithms, that can directly learn from pixels. Srinivas \textit{et al.} incorporated Contrastive Learning, into off-policy algorithms, to learn relevant features from image based inputs. Laskin \textit{et al.} investigated developing data efficient and generalizable algorithms, by introducing a generic data augmentation module for RL algorithms; \cite{laskin2020reinforcement, shang2021reinforcement, srinivas2020curl}.

While a lot of work has been devoted to extracting positional information from image inputs, very little investigation has been done on learning from temporal information. Shang \textit{et al.} performed experiments using DMControl (\cite{tassa2018deepmind}) to highlight the importance of temporal information in RL. They compared two Soft-Actor-Critic (SAC) RL algorithms, wherein one had access to pose and temporal information and the other only had access to pose. It was found that the former algorithm swiftly learned the optimal policy, while the latter failed to do so. Furthermore, a recurring heuristic used by many papers is to stack sequential observations together while inputting it to a neural network; \cite{mnih2015human}. This heuristic combines frames, without processing it and therefore can be considered analogous to early fusion; \cite{karpathy2014large}. Recently, Shang \textit{et al.} approached this as a video classification problem. This is a lucid approach, as considering a DRL state equivalent to a video, will help improve the prediction capabilities of the underlying neural network. Successful video recognition architectures use late fusion where all frames are processed, using neural networks, before they are combined together; \cite{shang2021reinforcement}, \cite{laskin2020reinforcement}. 

Moreover, a video stream consists of both spatial and temporal aspects. The former contains information about the video frame including objects and its surroundings, while the movement of the frame and its associated objects can be learned from the temporal portion; \cite{simonyan2014two}. While learning the spatial aspect is enough for image recognition, video recognition requires learning both spatial and temporal components. Enabling agents to extract temporal information from a given set of frames will result in the DRL agent making better Q-value predictions and therefore result in improved data efficiency. Furthermore, it will contribute to the agent understanding the differences between seemingly similar actions, such as opening and closing objects; \cite{lin2019tsm}, \cite{shang2021reinforcement}. 

There has been a plethora of work related to video recognition using 3D and standard 2D CNNs. 3D CNNs have the ability to simultaneously extract out spatial and temporal features from videos. However, they are computationally costly, which makes them hard to implement in real-time situations. Incorporating similar architectures with vision-based DRL exacerbates this problem, as many applications require fast predictions during training, and having latency is infeasible. Furthermore, the extra parameters could make the model more prone to overfitting without large amounts of data. This once again poses a roadblock to the development of sample efficient RL;\cite{lin2019tsm,tran2015learning, carreira2017quo}. 2D CNNs, although relatively efficient however fail to extract out temporal information;\cite{simonyan2014two}. 

\cite{amiranashvili2018motion} incorporated optical flow in their RL algorithm, although their technique required state variables in addition to pixel observations during training. Modeling temporal information in RL using simply pixel inputs was investigated by \cite{shang2021reinforcement}, and it brought a new approach to efficiently reducing sample complexity in reinforcement learning. We intend to further optimize this technique by leveraging recent work in the field of video action prediction and therefore propose the Temporal Shift Reinforcement Learning (TSRL) algorithm.

The contributions of our work \footnote{Our code is available at \url{https://anonymous.4open.science/r/TSM_RL-85F5/README.md}} are presented here:

1) We propose a plug and play architecture works with any generic vision-based DRL algorithms.

2) We augment a video recognition; \cite{lin2019tsm} that does not require any additional parameters to model temporal information in DRL.

\section{Related Work}
\subsection{Latent Flow} Simonyan \textit{et al.} investigated the use of optical flow techniques to perform video classification and was able to achieve SOTA performance by a significant amount over previous work in video classification. They developed a dual-stream architecture using ConvNets, consisting of spatial and temporal recognition components. The spatial stream was learned using a pre-trained ConvNet, wherein each frame was sent to the network as input. The input to the temporal stream was stacked optical flow displacement fields generated from consecutive frames. Movement among frames can be obtained from optical flow fields, thereby eliminating the need for the network to learn it. This technique achieved high accuracies without requiring a lot of data. More importantly, they established that training a temporal CNN using optical flow was a drastically better technique compared to training on a stacked bunch of images; \cite{simonyan2014two, karpathy2014large}. The downside of this algorithm is that it is computationally costly both during inference and training and therefore cannot be combined with RL algorithms; \cite{shang2021reinforcement}.

\subsection{Flow of Latents} Shang \textit{et al.} looked for a computationally feasible technique to integrate RL with optical flow. They were inspired by late fusion techniques; wherein every frame was run through a CNN before fusion was applied. Late fusion provides improved performance with lesser parameters and also allows multi-modal data \cite{jain2019learning, chebotar2017path}. They presented a structured late fusion architecture, wherein each image frame was encoded using a neural network. The encodings at each time step were subtracted from their prior, and this difference was fused with the latent encodings, which was then used by the RL algorithm. This technique was analogous to the work done by \cite{simonyan2014two}. The optical flow was approximated using the difference in encodings, which provided temporal information. The spatial component was obtained by encoding each of the frames. This technique provided the CNN with a necessary inductive bias. They chose Rainbow DQN, and RAD; \cite{laskin2020reinforcement} to be their base algorithm and found that it outperformed SOTA algorithms in performance and sample efficiency. Also, they showed that their algorithm reached optimal performance in state-based RL despite only being provided positional state information and no state velocity.

They also separately investigated encoding frames and then stacking the encodings together instead of the raw images. This technique yielded sub-par results, and the authors hypothesized that stacking high dimensional image frames would allow CNNs to learn temporal information. However, by stacking latent frames, the temporal information was lost and thereby causing the difference in results. 

\subsection{Temporal Shift Module}
While working with video model activations consisting of $N$ frames, $C$ channels and $H$ height and $W$ width, \\$A \in \mathbb{R}^{N X C X T X H X W}$, 2D CNNs don't consider the temporal dimension $T$ thereby ignoring it. \cite{lin2019tsm} addressed this by shifting channels, thereby mixing information from neighboring frames through the temporal dimension and referred to it as the Temporal Shift Module (TSM). Therefore the current frame contains information that was obtained from its surroundings. They leveraged the concepts of shifts and multiply-accumulate, which are the basic principles of a convolution operation. They extended it by shifting one step forward and backward along the temporal dimension. Furthermore, the multiply-accumulate was folded from the channel dimension to the temporal dimension. However, for online video recognition, only previous frames could be shifted forward and not the other way around. Therefore in such cases, a uni-directional TSM was implemented.

While this process doesn't require extra parameters, they found that this technique had drawbacks - 1) The data movement generated due to the shift strategy was not efficient and would increase the latency, especially since 5D activation of videos results in large memory usage. This implied that moving all channels would result in inference latency and large memory footprint on the hardware hosting the model. 2) Moving channels directly across the temporal dimension, referred to as in-place shift, would affect the accuracy of model since the spatial model is distorted. This is because the current channel would have some of its frames (or feature maps) shifted, and therefore, the 2D CNN would lose that information during the classification process. The authors obtained a 2.6 \% accuracy drop relative to their baseline; \cite{wang2016temporal} while naively shifting channels. The former issue was mitigated by shifting only a partial number of channels, thereby reducing the amount of data movement and latency incurred. For the latter problem, the TSM module was inserted within the residual branch of a Res-Net, thereby enabling the 2D CNN to learn spatial features without degrading. The authors claimed that this method, namely residual shift, allows the information present within the original activation to be retained after channel shifting due to identity mapping. Therefore, the TSM module is a simple modification to the 2D CNN. After encoding images, it shifts frames in the temporal dimension by +1, -1, and 0. However, shifting frames by -1, i.e. backward, is only possible for offline problems. For online image classification problems, the frames are moved +1; \cite{lin2019tsm}.

A major advantage of online TSM was that it enabled multi-level temporal fusion. Other online methods are generally limited to late and mid-level temporal fusion. The authors found multi-level temporal fusion to significantly influence temporal problems; \cite{zhou2018temporal,lin2019tsm, zolfaghari2018eco}.
 
\subsection{Prioritized Deep Q Network}
\cite{mnih2015human} combined Q Networks with CNNs in order to obtain an approximation of the Q values - 

$Q^{*}(s,a) = \max_{\pi}\mathbb{E}[r_t + \gamma r_{t+1} + \gamma^2 r_{t+2} + ... |s_t = s, a_t = a, \pi]$
\begin{figure*}[htp]
    \centering
    \includegraphics[width=14cm]{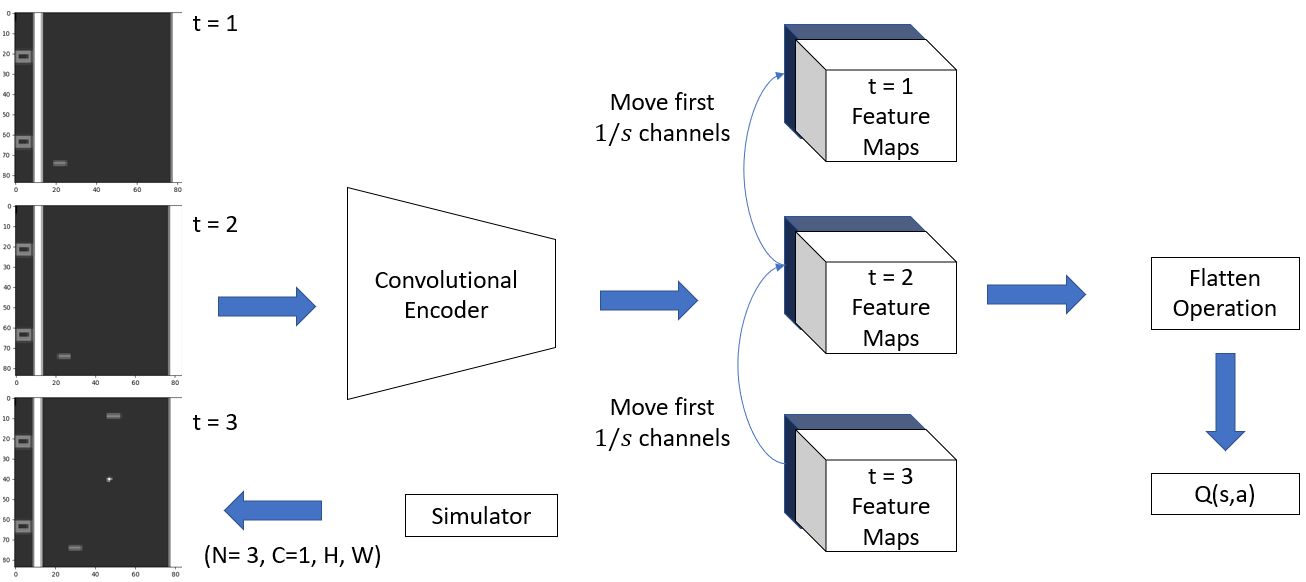}
    \caption{A schematic of Temporal Shift Reinforcement Learning algorithm}
    \label{fig:env}
\end{figure*}

The above expression maximizes the sum of discounted rewards for an agent following a policy, $\pi = P(a|s)$, $r$ using a discount factor, $\gamma$ during every time step $t$. It was known to be the first RL algorithm that could be integrated into various environments with raw pixels as inputs. They addressed the learning instabilities that RL presented when coupled with a deep neural network using a replay buffer and target network. They found that the sequential observations were highly correlated with each other and also that minimal changes to $Q$ would drastically affect the policy. The use of a replay buffer mitigated this issue by randomizing the data during the training process. This was done by storing the transitions as a tuple $(s,a_t,s_{t+1}, r{t+1})$ of state, action, next states and rewards within a cyclic buffer. This provided a two-fold benefit. The replay buffer reduced the number of environments needed for the agent to learn since the agent could always resample from the buffer. Furthermore, this reduces the variance during gradient descent since batches are sampled. The target network takes the weight from the current network but updates it only after a fixed duration of time. The target network's weights are then used to compute the TD error, which is the difference between the Q value and the TD target. If we use the parameters from the current network to estimate both these values, they'll become correlated and will result in instability. \cite{hasselt2010double} suggested using dual instead of single estimators to estimate the expected return since the latter led to over-estimated values and introduced the Double-Q learning algorithm (DDQN). A later investigation by \cite{van2016deep} showed that rather than learning a separate function, the target network could be used to obtain the estimate; \cite{mnih2015human, arulkumaran2017brief}. 

In addition, \cite{schaul2015prioritized} modified the experience replay process so that, instead of the conventional uniform sampling process, important samples were given a higher priority. The Prioritized Experience Replay (PER) technique was found to double the learning speed and also achieve SOTA scores on Atari games.

\section{Approach}

The motivation behind TSRL was to introduce an efficient algorithm that did not require any additional parameters, leveraging the benefits of multi-level temporal fusion. The architecture developed by \cite{lin2019tsm} for online Temporal Shift was modified and incorporated into a Double DQN with Prioritized Experience Replay (DDQN-PER). \cite{lin2019tsm} used a ResNet model for their experiments, however going with the conventional CNN models used by the vision RL community, we used a shallow three layer CNN. 

Also, we used in-place shift instead of residual shift wherein the channels were directly moved across the temporal dimension. We assumed that the accuracy improvements obtained, in predicting the Q values, while modeling the temporal aspect would compensate for the loss obtained due to spatial degradation. Furthermore, the online TSM algorithm \cite{lin2019tsm} cached the features in memory and then replaced it with those in the next time step. Our approach was to directly roll the features across time steps. 

Finally, the authors of the TSM paper found that the highest accuracy for the online model was obtained by shifting 1/8th of channels for each layer of the neural network. However, while testing our algorithm, we found that the best results were obtained when we shifted around 1/5 to 1/3 of our channels.   

A schematic of our algorithm has been given in Figure~\ref{fig:env} and a PyTorch based pseudocode for our algorithm has been presented here -
\begin{algorithm}
\caption{TSRL}\label{alg:cap}

\begin{minipage}{\linewidth}

\begin{lstlisting}
For each step t do
    For each convolution step do
        x = self.relu1(self.conv1(x))
        n,c,h,w = x.shape
        x = x.reshape(n//T, T, c, h, w)
        copy = torch.clone(x)
        x[:,:, :c//8, :, :] = torch.roll(x[:,:, :c//8, :, :],
        shifts = 1, dims = 1)
        x[:,0, :c//8, :, :] = copy[:,0, :c//8, :, :]
        z_t = FullyConnected(x)
    End For
End For
\end{lstlisting}

\end{minipage}
\end{algorithm}

\section{Experiments}
We tested our algorithm using OpenAI Gym Atari environments with visual images as input. An open-sourced implementation of DDQN (https://github.com/higgsfield/RL-Adventure) combined with PER was used. The images were converted to grayscale to speed up the learning process. To gauge the sample efficiency of TSRL we compared it with a generic DDQN-PER getting stacked images as input. Also, we used our own implementation of the algorithm developed by \cite{shang2021reinforcement} and referred to it as Flare, in order to compare against state of the art. The number of stacked images was kept equal to the timesteps considered by TSRL both for DDQN-PER and Flare. Also, all algorithms were run for 1.4M time steps using 5 different trials. The performance of the algorithm was gauged by averaging the trials and then summing over all rewards obtained; \cite{brockman2016openai, bellemare2013arcade, mott1996stella}.
\begin{figure}[!htp]
\begin{subfigure}{.5\textwidth}
  \centering
  \includegraphics[width=.8\linewidth]{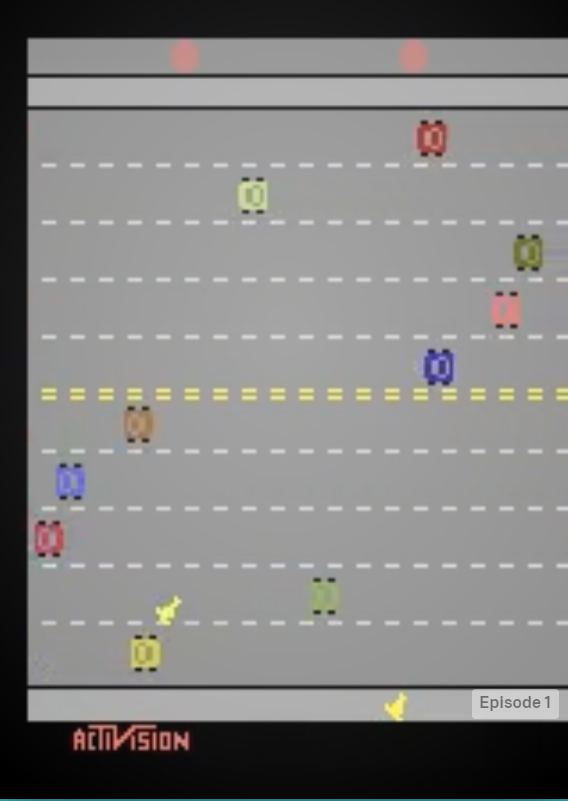}
  \caption{Freeway Atari Environment}
  \label{fig:sfig1}
\end{subfigure}%
\begin{subfigure}{.5\textwidth}
  \centering
  \includegraphics[width=.8\linewidth]{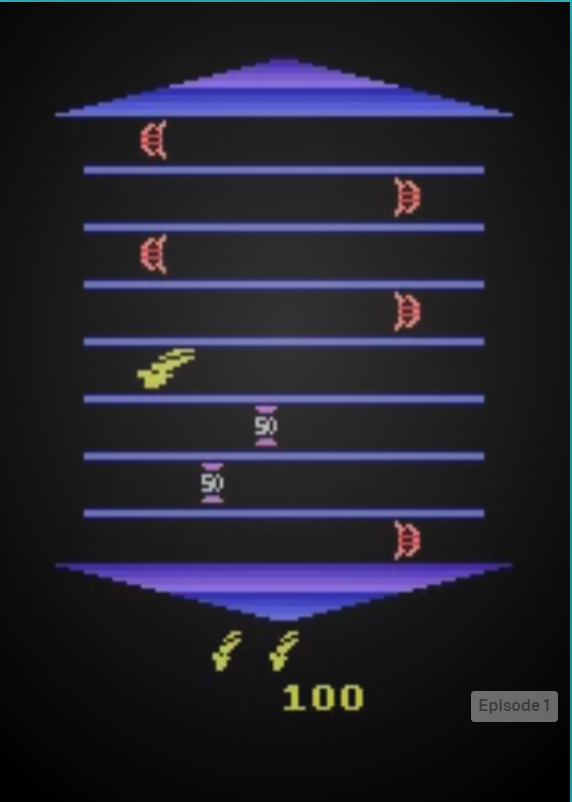}
  \caption{Asterix Atari Environment}
  \label{fig:sfig2}
\end{subfigure}
\begin{subfigure}{.5\textwidth}
  \centering
  \includegraphics[width=.8\linewidth]{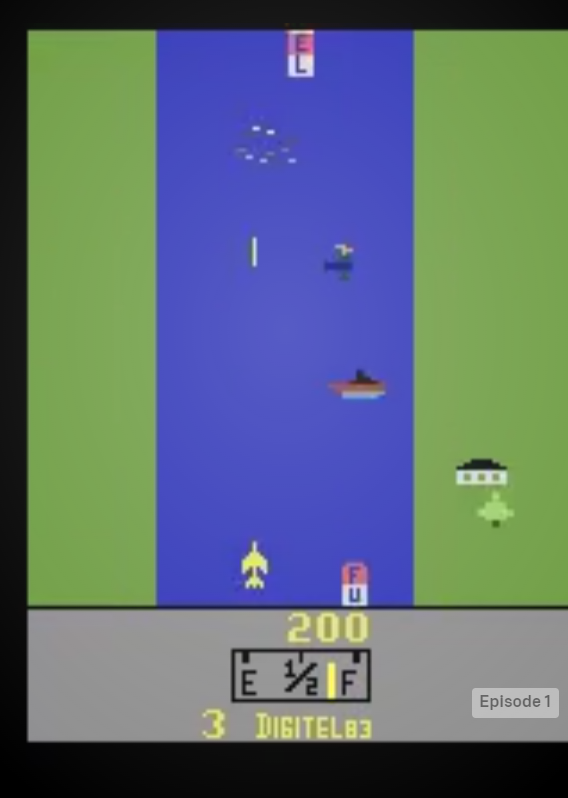}
  \caption{River Raid Atari Environment}
  \label{fig:sfig3}
\end{subfigure}
\begin{subfigure}{.5\textwidth}
  \centering
  \includegraphics[width=.8\linewidth]{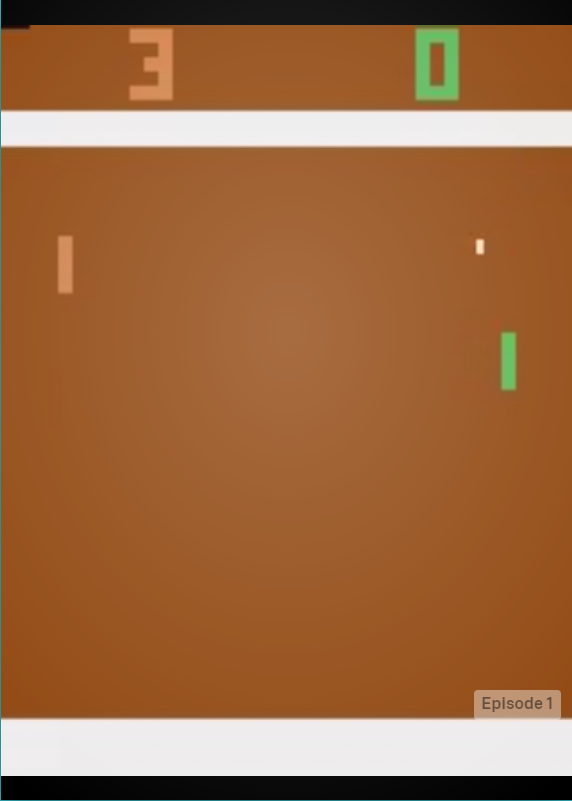}
  \caption{Pong Atari Environment}
  \label{fig:sfig4}
\end{subfigure}
\caption{OpenAI Gym environments used for training}
\label{fig:fig}
\end{figure}

\subsection{Results}
Table \ref{table:1} shows the sum of average rewards obtained across the five runs for each environment. The shift parameter, $s$ column, shows the ratio of channels that were shifted. For instance, if $s = 3$, then the first 1/3\ts{rd} channels would be shifted across the temporal dimension for every layer of the CNN.

\begin{figure}[!htp]
\begin{subfigure}{.5\textwidth}
  \centering
  \includegraphics[width=.8\linewidth]{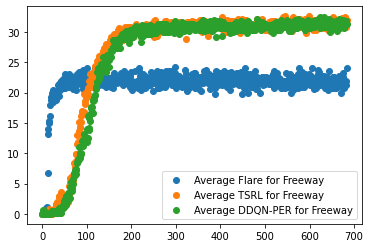}

  \label{fig:sfig1}
\end{subfigure}%
\begin{subfigure}{.5\textwidth}
  \centering
  \includegraphics[width=.8\linewidth]{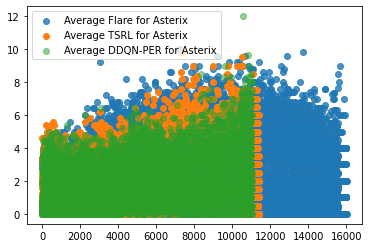}

  \label{fig:sfig2}
\end{subfigure}

\begin{subfigure}{.5\textwidth}
  \centering
  \includegraphics[width=.8\linewidth]{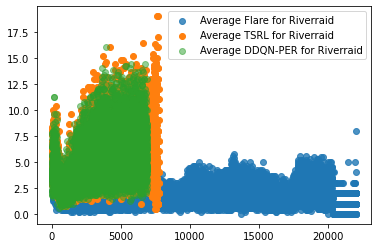}

  \label{fig:sfig3}
\end{subfigure}
\begin{subfigure}{.5\textwidth}
  \centering
  \includegraphics[width=.8\linewidth]{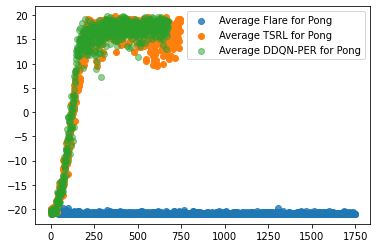}

  \label{fig:sfig4}
\end{subfigure}
\caption{Plots of episode vs reward for different Atari environments}
\label{fig:plots}
\end{figure}

Figure~\ref{fig:plots} shows the reward obtained per episode. In some cases, an algorithm may have large step sizes relatively early. This would lead to a lower number of episodes and vice versa. 

TSRL outperforms both DDQN-PER and Flare in all environments except Asterix, wherein it only defeats the DDQN-PER.

\begin{table}[h!]

\centering
\caption{Sum of average rewards obtained.}
\begin{tabular}{||c c c c c||} 

 \hline
 Environment & Shift & TSRL & DDQN-PER & FLARE\\ [0.5ex] 
 
 \hline\hline
 Freeway &3 & \textbf{18291.5} & 17807.6 & 14686.19\\ [1ex] 
\hline
 Asterix &5 & 22854.25 & 20702.0 & \textbf{33496.93}\\ [1ex] 
\hline
 Riverraid &5 & \textbf{41850.3} & 34849.2 & 34966.0\\ [1ex] 
\hline
Pong &5 & \textbf{7892.17} & 7221.80 & -36528.20\\ [1ex] 
\hline
\end{tabular}
\end{table}
\label{table:1}

\subsection{Discussion}
A major difference between our algorithm and other RL algorithms taking temporal aspects into account is that we provide multi-level temporal fusion. Most RL algorithms implement early fusion \cite{mnih2015human} and the recent ones \cite{amiranashvili2018motion, shang2021reinforcement} have experimented with late fusion. However, our approach enables RL to have temporal fusion across all levels. This type of fusion was found to significantly help difficult temporal modeling problems \cite{lin2019tsm}. 

It is interesting to note that instead of a single shift hyperparameter being optimal for all tasks, it varies across environments. We hypothesize that this is caused due to the trade-off between spatial and temporal learning. Some environments might not require a higher number of feature maps and therefore could work with a lower shift hyperparameter. This would permit a larger number of channels to be moved, leading to improved temporal learning. However, this might not be the case in complicated environments, and such situations might require the shift hyperparameter to be higher.

Finally, we see that TSRL is able to beat the baseline and SOTA for almost all the environments.  \footnote{We used our own implementation of the Flare algorithm.} Since Flare concatenates latent flow with features, we feel that this increases the number of parameters and, therefore, the relative training time compared to TSRL. Furthermore, the latent flow is obtained by subtracting the current frame from the immediately preceding frame while ignoring the frames before that. This might not provide much information in situations when the difference between immediate frames is minute. This problem is mitigated by the multi-level fusion abilities of our algorithm.
\section{Conclusions}
We present a facile shifting technique for learning temporal features in DRL problems without the requirement of additional parameters. After testing our algorithm on OpenAI Atari environments, we find that our algorithm outperforms the commonly used frame-stacking heuristic.

A major drawback of our algorithm is the requirement to find a suitable shift hyperparameter. Future work could include either learning the optimal value of this hyperparameter online or changing how the shift is performed (such as residual shift \cite{lin2019tsm}) so that the spatial features aren't disturbed.

\bibliography{iclr2021_conference}
\bibliographystyle{iclr2021_conference}

\end{document}